\documentclass[10pt,twocolumn,letterpaper]{article}

\usepackage{cvpr}
\usepackage{times}
\usepackage{epsfig}
\usepackage{graphicx}
\usepackage{amsmath}
\usepackage{amssymb}

\usepackage[pagebackref=true,breaklinks=true,letterpaper=true,colorlinks,bookmarks=false]{hyperref}

\cvprfinalcopy 

\def\cvprPaperID{****} 

\ifcvprfinal\fi
\begin{document}

\title{Multi-modal Transfer Learning for Dynamic Facial Emotion Recognition in the Wild}

\author{Ezra Engel\\
{\tt\small eengel9@gatech.edu}\\
Georgia Institute of Technology\\
\and
Chris Hudy\\
{\tt\small chudy3@gatech.edu}\\
Georgia Institute of Technology\\
\and
Lishan Li\\
{\tt\small lli680@gatech.edu}\\
Georgia Institute of Technology\\
\and
Robert Schleusner\\
{\tt\small rschleusner@gatech.edu}\\
Georgia Institute of Technology\\
}

\maketitle

\begin{abstract}
   Facial expression recognition (FER) is a subset of computer vision with important applications for human-computer-interaction, healthcare, and customer service. FER represents a challenging problem-space because accurate classification requires a model to differentiate between subtle changes in facial features. In this paper, we examine the use of multi-modal transfer learning to improve performance on a challenging video-based FER dataset, Dynamic Facial Expression in-the-Wild (DFEW). Using a combination of pretrained ResNets, OpenPose, and OmniVec networks, we explore the impact of cross-temporal, multi-modal features on classification accuracy. Ultimately, we find that these finely-tuned multi-modal feature generators modestly improve accuracy of our transformer-based classification model. 
\end{abstract}

\section{Introduction/Background/Motivation}

Facial emotion recognition (FER) is a challenging subfield of computer vision with important applications to human-computer interaction. The human face is an essential mechanism through which people communicate information, and current literature suggests a large fraction human communication consists of body language, facial expression, and vocal tone \cite{duncan1969nonverbal}\cite{hall2019nonverbal}. As computers become an increasingly important feature of our daily lives and work, it is essential that we improve their ability to recognize and respond to these nonverbal communication signals.

Automatic computer recognition of human emotion has broad applications to a wide range of human-computer interaction systems including customer service, driver-fatigue detection, computational therapy, video-game design, and many others \cite{canal2022survey}\cite{chowdary2023deep}. However, facial expression recognition outside of laboratory settings remains an extremely difficult task \cite{farzaneh2021facial}. Many current models focus on only a single mode of communication (text, facial expression, and audio-sentiment) \cite{canal2022survey}. We aim to use a suite of models to incorporate information from a variety of these modalities and improve performance on real-life applications of human sentiment classification.

Many current and historical models focus on static facial expression recognition. In these models, a series of still, laboratory-controlled image of faces are classified into seven emotion categories: happiness, sadness, neutral, anger, surprise, disgust, and fear \cite{canal2022survey}. The laboratory task uses frontal facial images under standardized lighting conditions which greatly simplifies the task. Accurate FER in unconstrained real-world settings (in-the-wild) remains extremely difficult.

Approaches that emphasize in-the-wild applications still generally focus on using a single modality to classify a facial expression. These consist of both static (image) and dynamic (video) classification tasks. Previous research has seen some success in applying 3D convolutional networks to sequential image data \cite{li2017multimodal}\cite{hasani2017facial}, but the overall performance of single-mode models remains inadequate for many real-life applications \cite{canal2022survey}.

Multimodal approaches have gained significant popularity in recent years as deep learning facilitate a shift away from hand-craft features towards learned representations. Much current work has demonstrated the usefulness of multimodal sentiment data in improving model-accuracy for in-the-wild classification tasks.

Video-based FER datasets that include audio and other contextual cues facilitate a move beyond static imagery. In this work, we target the Dynamic Facial Expressions in the Wild (DFEW) dataset--a large scale FER dataset consisting of 16,372 video clips from over 1,500 movies, each annotated with intensities for seven primary emotion categories. DFEW provides short, unconstrained video clips which makes it ideal for spatiotemporal and multimodal approaches.

We aim to apply many of these findings to a relatively new dataset which focuses on dynamics FER in-the-wild, DFEW \cite{jiang2020dfew}. While most other currently available in-the-wild FER datasets consist of still images, the DFEW dataset consists of 16,372 unconstrained facial expression video clips with robust annotations.

In this paper, we propose a multimodal FER model that leverages transfer learning from SOTA pretrained models in three modalities--visual, audio, and pose. Our architecture integrates a ResNet-18 CNN (for facial frames), an OpenPose network (for body posture), and a Wav2Vec2 model (for speech audio) to extract complementary features from each clip. These features are then combined using a transformer-based sequence model to capture temporal dependencies before final classification in a shallow dense network. Ultimately, this approach yields improved performance on dynamic in-the-wild data, outperforms the original DFEW baseline, and approaches the accuracy of recent SOTA architectures.

\section{Approach}

\begin{figure*}
\begin{center}
\includegraphics[width=0.65\linewidth]{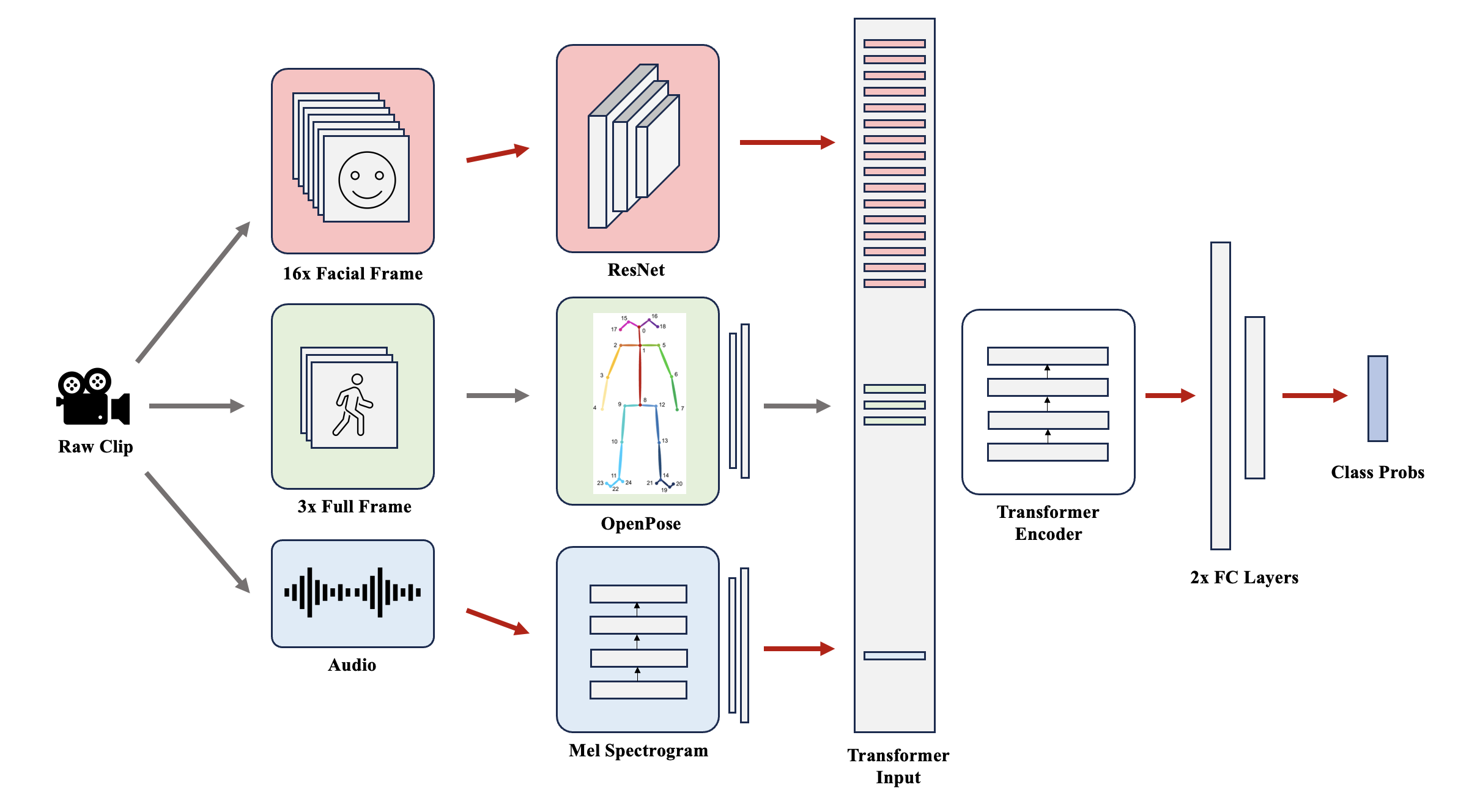}
\end{center}
   \caption{\textbf{Original Multi-modal FER model architecture.} We combine three pretrained models to extract facial features, body-pose data, and audio-sentiment from the processed and raw clips. Then, we leverage a multilayer transformer encoder to learn spatiotemporal dependencies and relationships to improve FER classification accuracy. Ultimately, we opted to connect the Mel Spectrogram and OpenPose features directly to the FC layers after poor initial results.}
\label{fig:architecture}
\end{figure*}

In contrast to many prior FER approaches which perform feature fusion early in the architecture, our design leverages modular pretratined feature extractors for each modality and defer their integration to a deeper layers. By using separate pretrained models for visual, auditory, and pose cues, we can exploit representations learned from large-scale unimodal datasets like ImageNet. Decision-level fusion also offers training flexibility, since each modality can be trained in parallel with limited interference. We hypothesize that this architecture facilitates the learning of cross-modal relationships, since each input stream is first projected into a pipelined latent feature space.

We break the FER classification task into four steps: (1) image/audio pre-processing, (2) feature extraction, (3) cross-modal and temporal integration, and (4) the decision network. Specifically, we fine-tune pretrained feature extractors in an end-to-end manner while training a multi-layer transformer encoder to integrate features across modalities and temporal space. A full diagram of our architecture is shown in Figure \ref{fig:architecture}.

\subsection{Image Pre-Processing}

The creators of the DFEW dataset include both raw clips and pre-processed frames. The original authors used OpenCV to extract image frames from the raw clips, and then applied face++ API to extract face region images and facial landmarks \cite{face++}. They then normalized the extracted faces using SeetaFace \cite{liu2017viplfacenet}. Finally, the authors standardize the sequence length by interpolating between the processed sequence images to create 16 pre-processed frames for each clip \cite{zhou2011towards}\cite{zhou2013compact}.

\subsection{Audio Pre-Processing}

For the audio data, we utilize a pretrained Wave2Vec2 model to extract high-dimensional embeddings from the raw audio clips. Specifically, we process the raw audio through the Wave2Vec2 framework, which generates a 1024-dimensional feature vector \cite{wav2vec2_framework}. The extracted embeddings are normalized to ensure consistent scaling. Subsequently, we reduce the dimensionality of the normalized embeddings by applying a series of linear layers interleaved with batch normalization layers, first reducing to 512 dimensions and then to 128 dimensions \cite{wav2vec2_feature_extractor, wav2vec2_hubert_emotion_recognition}. The final 128-dimensional feature vector represents the processed audio and is prepared for integration with the video and OpenPose features in subsequent stages.

\subsection{Feature Extraction}

We use three pretrained models to extract spatiotemporal features from the raw clips, the preprocessed facial frames, and the audio data. Two of these models (ResNet \cite{he2015resnet} and Wave2Vec2 \cite{wav2vec2_framework}) are fine-tuned while training the transformer and decision layers.

We use the Torch Vision implementation of ResNet-18 pretrained on ImageNet as a base-model for facial feature extraction \cite{pytorch}. This pipeline is show in the top branch of Figure 1. First we strip the final fully-connected layer off of the pretrained model (512x1000). Then, for each of the 16 pre-processed frames we generate a length-512 vector which captures meaningful abstracted features from the frame. These vectors, are concatenated into a 16-length sequence which is then summed with learnable position embeddings and feed the multi-layer transformer encoder.

We extract body-language information using a pretrained PyTorch implementation of OpenPose \cite{cao2019openpose}. This model processes three full-body frames extracted from the raw video clips, generating keypoint data for 17 various joints and features as a tensor of shape (17, 3) per frame. We drop the third value, representing a confidence score consistently equal to 1, reducing the tensor to (17, 2). These keypoints are flattened and passed through a fully connected layer, transforming them into a higher-dimensional representation of shape (512, 3), where each column corresponds to one frame. Finally, these features are concatenated with the video-based embeddings, resulting in a combined feature sequence of length 19.

Finally, we use a pretrained audio model (e.g., Wave2Vec2 \cite{wav2vec2_feature_extractor, wav2vec2_hubert_emotion_recognition}) to extract a 1024-dimensional embedding from the audio data. After normalizing this embedding, we apply a series of linear and batch normalization layers to reduce its dimensionality first to 512 and then to 128. This resulting 128-dimensional audio feature vector is then concatenated with the video and OpenPose features before being passed to the sequence transformer.

\subsection{Transformer Layer}
Our 20-length sequential feature vector is the direct input to a multi-layer transformer encoder using the built-in PyTorch implementation \cite{pytorch}\cite{vaswani2017attention}. We use an encoder with 6 layers, 8 heads, and a feed-forward size of 2048. The goal of the transformer encoder is to take out sequential multi-modal inputs and use them to produce a new set of features which effectively capture the cross-temporal and cross-modal relationships between our extracted features. The output of the encoder is then flattened and passed through our decision network.

\subsection{Decision Network}

The decision network consists of two fully connected layers with a final output size of 7. The output of the final affine transformation is passed to a log-softmax function to calculate normalized class scores. The DFEW dataset includes both one-hot labeled class attributions \textit{and} individual class scores for each clip \cite{jiang2020dfew}. We take some inspiration from knowledge distillation and choose to conduct training directly on the class scores (as opposed to the one-hot label). To this effect, we use a Kullback-Leibler loss functions between predicted class probabilities and the ground-truth class scores. Our hope is that by including ambiguous and nuanced scores, the model is better able to learn relationships and correlations between subtle differences in our extracted features. We expect this to improve overall model performance.

\subsection{Evaluation Metric}

To facilitate direct comparison with the DFEW authors' benchmark and to deal with class imbalances, we use Weighted Average Recall (WAR) as our evaluation metric for the classification task \cite{jiang2020dfew}. The WAR is simply the accuracy of the classifier. The original authors also use an Unweighted Average Recall decribed in Equation \ref{eq:uar}, where $n$ is the number of classes, $tp_k$ is true positives, $tn_k$ is true negatives, and $T$ is total predictions. (The subscript $k$ denotes per class predictions).

\begin{equation}
    UAR = \sum^{n}_{k=1}\frac{1}{n}\frac{tp_k + tn_k}{T}
\label{eq:uar}
\end{equation}

We constrain our main analysis to WAR only, but include precision recall curves in Section \ref{sec: results}. When calculating the metrics, we first choose to ignore all samples for which none of the class scores is greater than 6 (the same threshold used in the original DFEW paper). This threshold serves to eliminate sufficiently ambiguous samples from a calculation of WAR with the understanding that these examples do not cleanly map to single-class labels.

\subsection{Approach Discussion}

Much previous work in multimodal FER focuses primarily on applying modality fusion within the feature extraction part of the network. While this approach has achieved relatively high performance, feature-fusion can make it difficult to leverage pretraining on larger corpuses and also complicates the use of self-supervised or semi-supervised techniques for performance improvements.

By combining relatively modular feature extractors with a robust transformer architecture we hope to take advantage of both multi-modal information \textit{and} pretraining on larger datasets and semi-supervised tasks. By effectively integrating audio and gesture information into the spatiotemporal decision architecture (the transformer and fully connected layers), we hope to observe signficant improvements over a baseline ResNet/Transformer model which uses only the 16 preprocessed facial frames.

FER in the wild is a notoriously difficult task (with the DFEW authors establishing a UAR benchmark of around 45.35\%). Given the difficulty of the task and a relatively high-capacity model we anticipated difficulties with both training and overfitting. For these reasons, we anticipated having to perform significant hyperparameter tuning (ResNet depth, transformer depth, learning rate, scheduler, dropout) to achieve reasonable performance on the evaluation sets.

Our initial ResNet model (single mode), performed significantly worse than the DFEW benchmarks. However, during hyperparameter tuning we found that the performance could be significantly improved by lowering the learning rate. We also found that using multimodal features as inputs to the transformer badly degraded overall model performance. Bypassing the transformer layer and including these multimodal inputs as direct input to the fully connected decision layers had a small, but noticeable impact on overall performance. These experiments are further discussed in the next section.

\section{Results and Discussion}

Our final model was a result of an empirically driven iterative design process. During development and testing, we found that (1) the model was highly sensitive to choice in our hyperparameters and learning rate, and (2) including cross-modal feature vectors as transformer inputs degraded the performance of the model.

\subsection{Hyperparameter Experiments}

The backbone model consists of pretrained ResNet18 layers used as a feature extractor, paired with a transformer model to learn how to focus on changes across consecutive facial frames (16 frames) in order to identify the core emotion expressed in each clip. The first experiment involved training the ResNet-Transformer hybrid model end-to-end. The ultimate goal for model tuning is to achieve relatively high training accuracy while keeping the difference between training and testing metrics small, to mitigate high variance and overfitting.

We trained the baseline model using two different loss functions: mean-squared error and Kullback-Leibler divergence. We decided to explore both due to the ambiguity of the class labels. One potential interpretation of the 7-length annotation vector is a list of normalized scores for each individual class. For the case in which these labels represent normalized scores, mean squared error would constitute an appropriate metric. Another valid interpretation of the labels is as a discrete probability distribution over the 7 emotion classes. In this case, the KL-divergence is a more appropriate metric. For this reason, we explore the performance of both loss functions, and ultimately conclude that they perform similarly. These experiments are detailed in the following sections.

\subsubsection{Mean Squared Error}
Initially, both models showed limited progress, with training accuracy stagnating between 26\% and 35\% over 10 epochs, indicating that the models struggled to learn from the training data. Upon further analysis, we discovered a class imbalance issue, with the largest class containing 4,209 samples and the smallest class having only 145 samples. After applying stratified data splitting based on class distribution and normalization on training data, the model began to learn incrementally.

Figure 2 shows the results of the first 10 epochs of the MSE model, where we observed an increasing gap between training and validation loss, while both training and validation accuracies steadily increased, indicating effective learning. To further enhance performance, we trained the model for an additional 5 epochs with L2 regularization applied to prevent overfitting. However, the validation accuracy did not significantly improve after these extra epochs. 

The parameter tuning process primarily focused on adjusting the learning rate and weight decay in the Adam optimizer. The best-performing MSE model, based on validation loss, achieved 80.13\% training accuracy and 70.37\% validation accuracy, delivering consistent results during training while keeping over fitting under control.

\begin{figure}[htp]
    \centering
    \includegraphics[width=8cm]{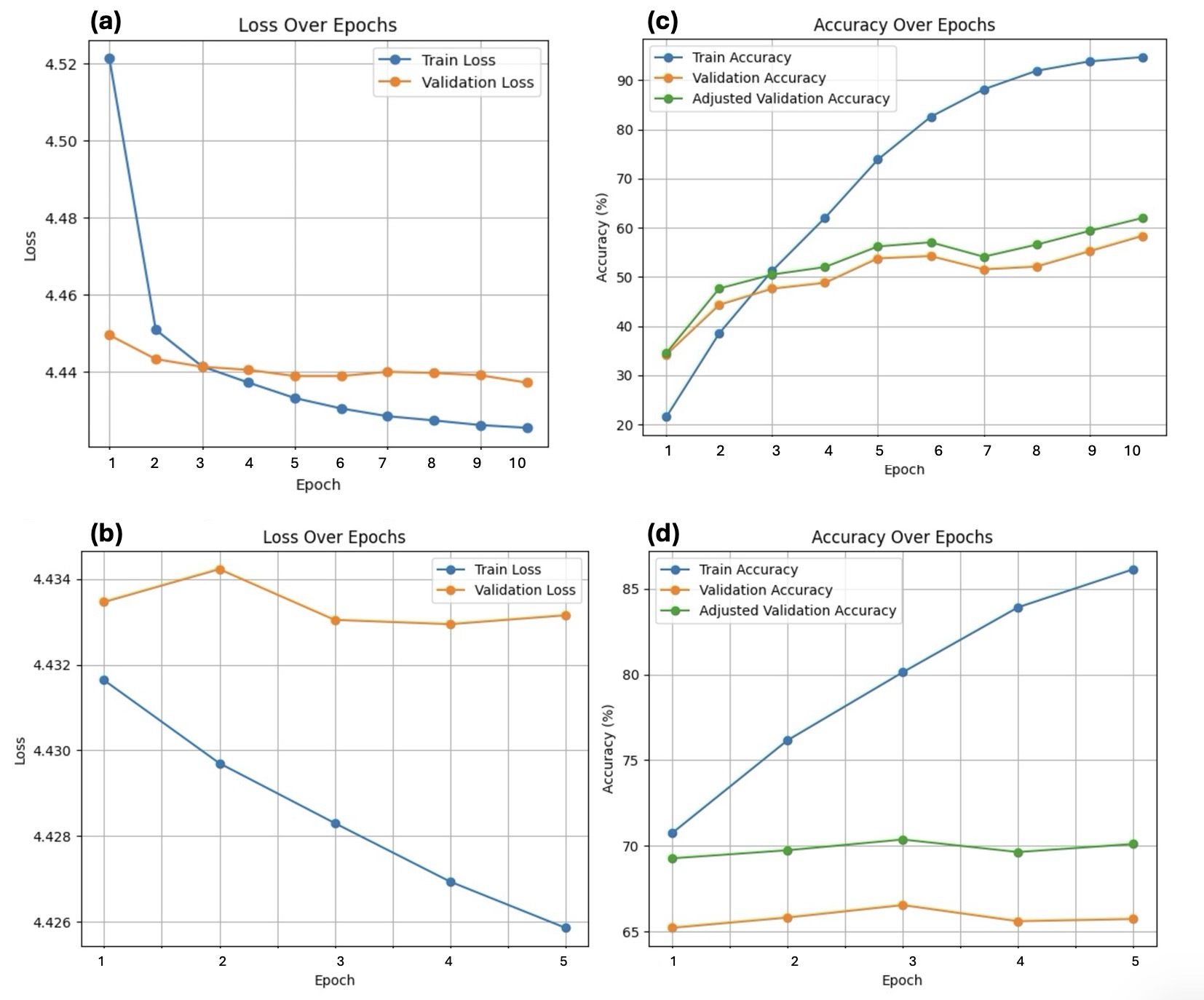}
    \caption{Model performance of the emotion classifier trained using Mean Squared Error (MSE) loss. Plots (a) and (c) show results from the first 10 epochs without regularization, while plots (b) and (d) represent performance with a regularization term (\texttt{weight\_decay})  set to 0.0001.}
    \label{fig:Model-Performance-with-MSE-loss}
\end{figure}

\subsubsection{Kullback-Leibler Divergence}
The training of the KLD model began with default parameter values: a learning rate of 0.0001, the Adam optimizer, and KL Divergence as the loss function. During the first 10 epochs, the training loss consistently decreased, while the validation loss began to increase. Despite this divergence, there were signs of improvement in validation accuracy illustrated in Figure 3.

To address the class imbalance issue, we switched to cross-entropy loss with class weights, but no significant improvement was observed after an additional 5 epochs. To reduce the gap between training and validation performance, regularization was applied with a weight decay of 0.0001, but this also failed to yield measurable improvements.

We were able to further improve accuracy of the model using an initial learning rate of 0.0001 combined with a scheduler to decrease learning rate on plateau. This final KLD model achieved a training accuracy of 98.67\% (due to weight decay) and a validation accuracy of 72.40\%.

\begin{figure}[htp]
    \centering
    \includegraphics[width=8cm]{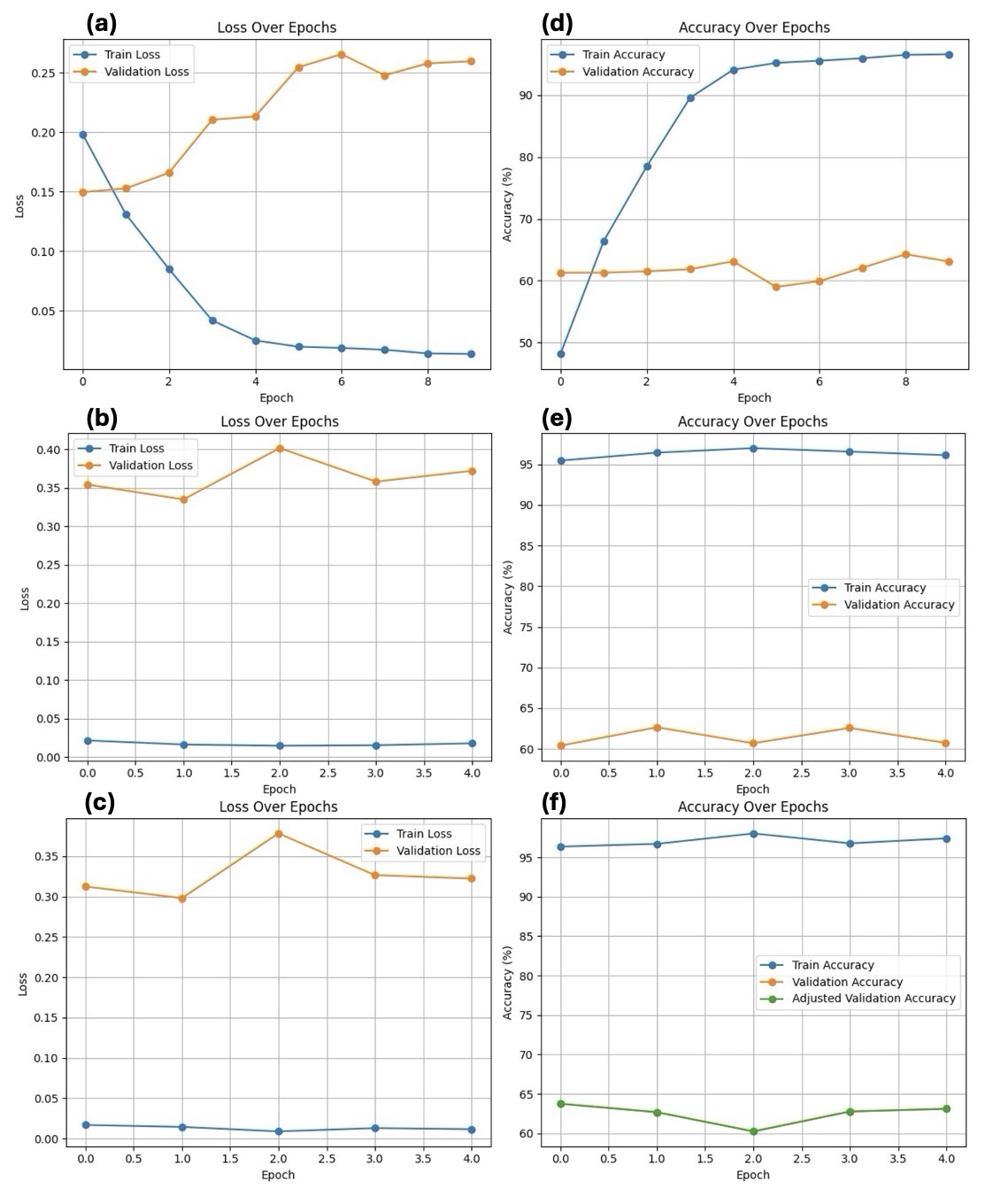}
    \caption{Model performance of the emotion classifier trained with KL Divergence (KL) loss is illustrated. Plots (a) and (d) show the results from the first 10 epochs using KL loss. Plots (b) and (e) depict performance after switching from KL loss to Cross Entropy loss, weighted by class distribution. Finally, plots (c) and (f) demonstrate performance when a regularization term (\texttt{weight\_decay}) is applied with a value of 0.0001.}
    \label{fig:Confusion Matrixes with KL and Cross Entropy loss}
\end{figure}

\subsubsection{KLD Model with ResNet (Frozen Layers)}
\begin{figure}[htp]
    \centering
    \includegraphics[width=8cm]{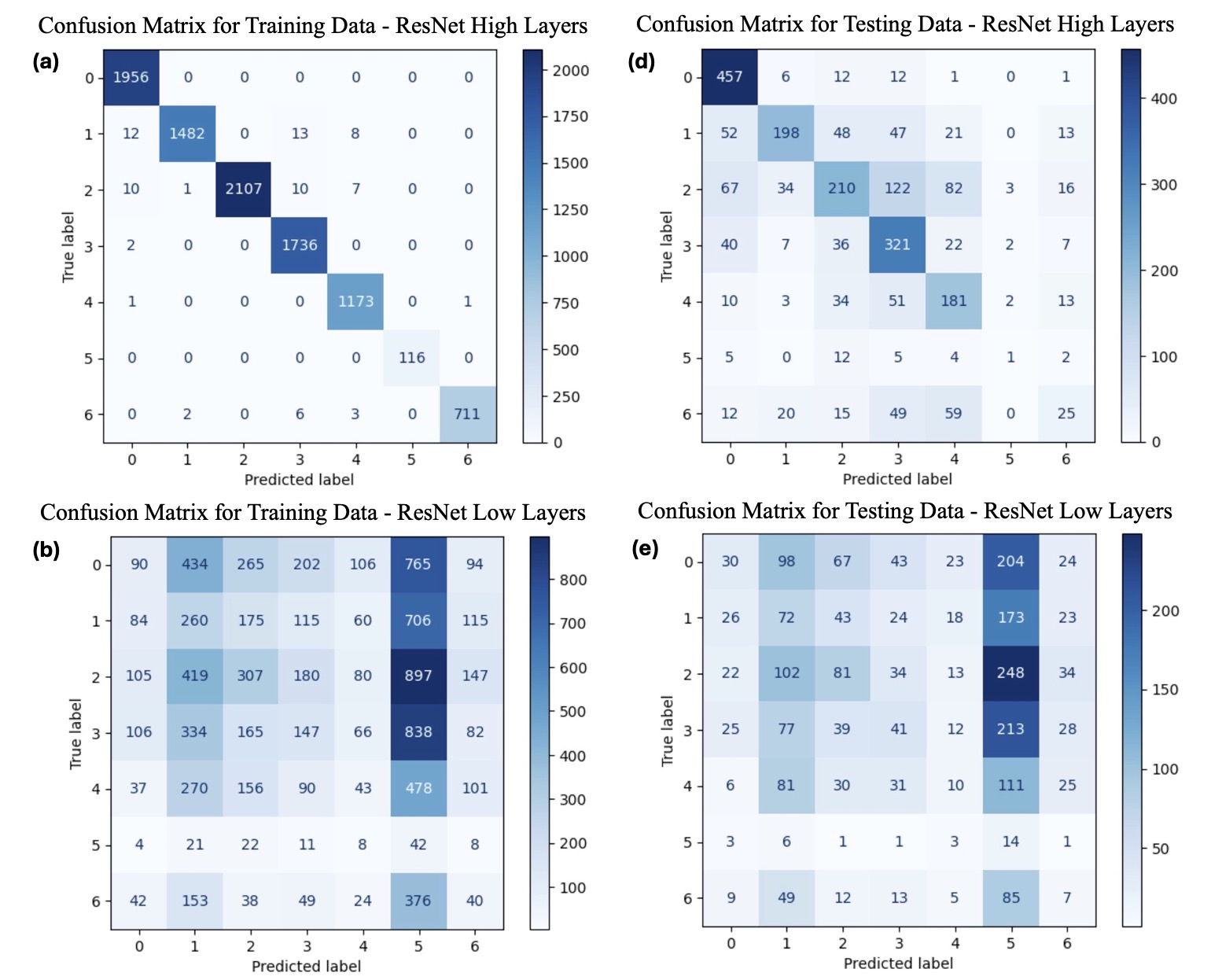}
    \caption{Confusion matrices for vlassification model trained with lower or higher ResNet layers.}
    \label{fig:Confusion Matrixes with KL and Cross Entropy loss}
\end{figure}

The initial approach of passing images through a pretrained ResNet model to extract facial features aimed to enhance training efficiency. The backbone model was trained end-to-end, combining ResNet with a transformer architecture. As shown in previous results, this approach demonstrated strong capabilities in differentiating human emotions. ResNet, pretrained on millions of images from the ImageNet dataset with 1,000 object classes, serves as an effective feature extractor. However, it remains unclear whether we need to retrain the entire ResNet model or if fine-tuning only partial layers would be sufficient. To address this, we conducted a comparison analysis.

The research paper on how CNN model learns visual features \cite{Zeiler2014Visualizing} suggests that the lower layers of ResNet focus on learning basic features like colors, edges, and lines, while the higher layers capture more abstract, task-specific features. Lower-level features are general and transferable, while higher-level features are tailored for specific tasks. Figure 4 illustrates this: freezing all ResNet layers except the last three high-level layers (high-level ResNet) resulted in performance comparable to the end-to-end model. In contrast, freezing all but the first three lower-level layers (low-level ResNet) led to poor performance, with predictions biased toward a single class out of the eight emotion categories.

These results are consistent with expectations and lead to a key conclusion: when using a pretrained model for a downstream task, focusing training on higher-level layers typically yields better performance. Furthermore, the number of trainable parameters for the end-to-end model, high-level ResNet, and low-level ResNet are 21,691,463, 21,008,391, and 10,524,487, respectively. This indicates an approximate 1\% reduction in computational resources while maintaining similar model performance. Besides, the significant difference in trainable parameters between the high-level and low-level ResNet models helps explain the failure of the low-level ResNet.

\subsection{Multimodal Experiments}

Our initial attempts to use OpenPose and Wave2Vec2 features as transformer inputs failed badly. The inclusion of these features actively degraded model performance. While a well-tuned ResNet-only model might achieve close to 70\% validation accuracy, multimodal models would consistently fail to surpass 30\% validation accuracy with similarly poor performance on the training set.

We hypothesize that the initial misalignment between these multimodal features was so large that the self-attention heads within the transformer architecture were unable to identify meaningful relationships between the inputs. We had initially supposed that through back-propagation, the model might be able to map the ResNet features, OpenPose features, and Wave2Vec2 into the same abstracted feature space. However, after repeated experiments, it became clear that attempting to fine-tune these models towards a coherent features space through the transformer layers was unlikely to be successful.

Instead, we opted to bypass the transformer entirely with the multimodal inputs. Instead of connecting these features to the transformer, we connect them to the final fully connected network. This avoids many of the issues with trying to back-propagate dissimilar feature spaces through the self-attention mechanism. Once we shifted to this new architecture, we saw small performance improvements of the model. A full summary of results is shown in Table \ref{tab:WAR for various models}.

\subsection{Results}
\label{sec: results}

Figure \ref{fig:Confusion Matrixes with KL and Cross Entropy loss} summarized the overall performance of the ResNet-only KL divergence model. The model achieves relatively high overall classification accuracy but struggles on less common emotions such as disgust and fear. This is a common problem observed in the literature and has two primary causes. First, in-the-wild datasets tend to heavily preference happy, sad, angry, and neutral faces, since these phases tend to be the most common emotions across non-laboratory images.

\begin{figure}[htp]
    \centering
    \includegraphics[width=8cm]{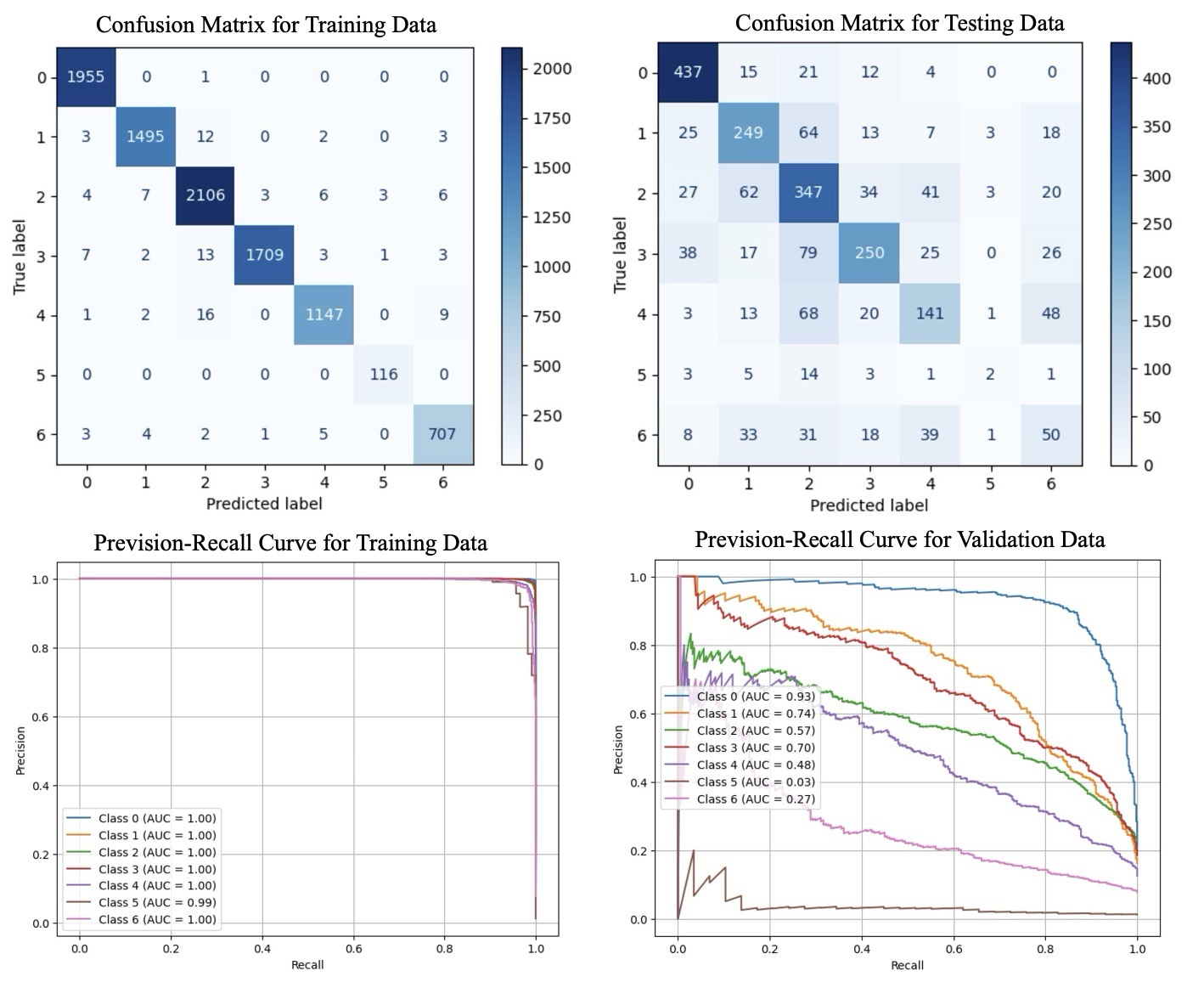}
    \caption{Confusion matrices and precision-recall curves for the training and testing datasets of the emotion classifier trained with KL loss.}
    \label{fig:Confusion Matrixes with KL and Cross Entropy loss}
\end{figure}

\begin{figure}[htp]
    \centering
    \includegraphics[width=8cm]{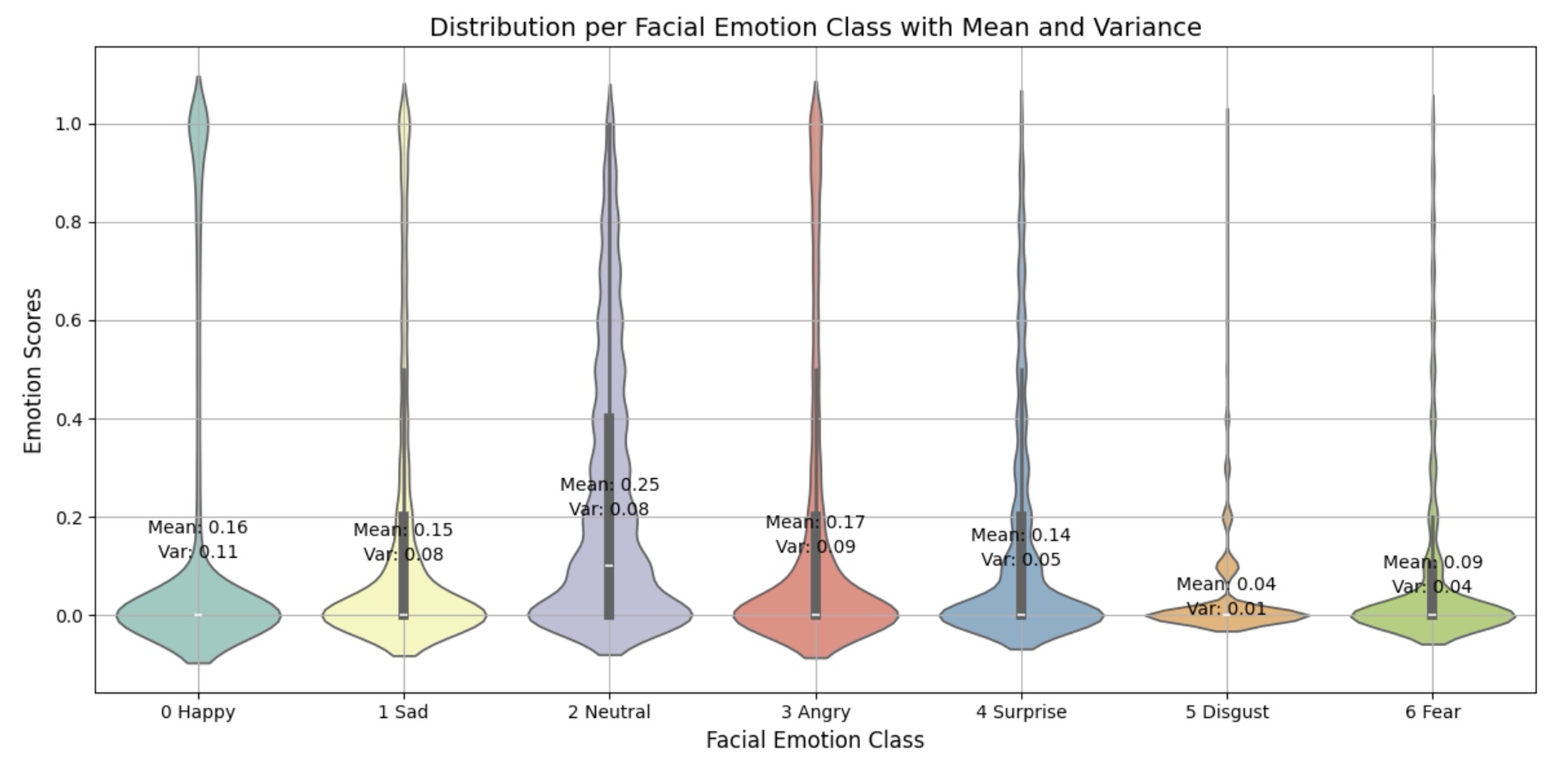}
    \caption{Box plots of data distribution per emotion class.}
    \label{fig:Distribution of Emotion Scores}
\end{figure}

 Second, disgust and fear tend to have more subtle effects on facial landmarks than do other major emotions (smiles, for instance, are extremely prominent and recognizable). Figure \ref{fig:Distribution of Emotion Scores} shows large skews in the distributions of most emotion classes. Specifically, the ``Happy", ``Sad", and ``Angry" classes tend to be labeled with higher scores, suggesting that these emotions are often associated with stronger facial expressions. This also makes them easier to identify, as illustrated in Figure \ref{fig:Confusion Matrixes with KL and Cross Entropy loss}, compared to the other four emotion types. This is consistent with findings in previous literature \cite{lek2023academic}.

Table \ref{tab:WAR for various models} shows the WAR for a selection of our ResNet-Transformer architectures and models from the literature. As discussed, when the OpenPose and Wave2Vec2 features are connected to the transformer layer, they degrade model performance by confusing the self attention mechanism. In contrast, when the multimodal features are connected directly to the FC layers, we see a modest improvement in classification accuracy.

Our final model outperforms the original DFEW authors' implementation of Expression-Clustered Spatiotemporal Feature Learning (EC-STFL) and performs relatively well when compared to other benchmarks.

\begin{table}
    \centering
    \begin{tabular}{|l|c|}
         \hline
         \textbf{Model} & \textbf{WAR} \\
         \hline
         ResNet-Only & 72.40\%\\
         ResNet+OpenPose$^{T}$ & 25.74\%\\
         ResNet+Wav2Vec2$^{T}$ & 26.87\%\\
         Full Multimodal$^{FC}$ & \textbf{72.97\%}\\
         EC-STFL\cite{jiang2020dfew} & 56.51\% \\
         M3-DFEL\cite{wang2023rethinking} & 69.25\% \\
         MMA-DFER\cite{chumachenko2024mma} (SOTA) & \textbf{77.51\%} \\
         \hline
    \end{tabular}
    \vspace{0.2cm}
    \caption{WAR for an assortment of both our model architectures and other benchmarks from the literature. Superscript $T$ and $FC$ denote multimodal features connected to the transformer layer and fully connected layer, respectively. EC-STFL is the DFEW authors' original benchmark.}
    \label{tab:WAR for various models}
\end{table}

The performance of the ResNet-Transformer model that we propose in this paper might be further improved with some additional refinements. The alignment of multimodal feature spaces is a known problem within multimodal learning problems \cite{xu2019learning}. One potential solution (and area for future work) would be to first train a shallow fully connected network to map these different features into the same feature space. This step-by-step training might avoid the issues resulting from dissimilar inputs to the self-attention mechanism.

{\small
\bibliographystyle{ieee_fullname}
\bibliography{egbib}
}

\clearpage
\end{document}